\documentclass[a4paper, 10pt, conference]{ieeeconf}      % Use this line for a4
\usepackage{FG2020}
\usepackage{times}
\usepackage{epsfig}
\usepackage{graphicx}
\usepackage{amsmath}
\usepackage{amssymb}
\usepackage{subcaption}
\usepackage[normalem]{ulem}

\let\labelindent\relax
\usepackage{enumitem}

\newcommand{\parahead}[1]{\textbf{#1}\ }

\FGfinalcopy % *** Uncomment this line for the final submission

\IEEEoverridecommandlockouts                              % This command is only
\def\FGPaperID{33} % *** Enter the FG2020 Paper ID here

\title{\LARGE \bf
Generative Model-Based Loss to the Rescue: A Method to Overcome Annotation Errors for Depth-Based Hand Pose Estimation
}

\author{\parbox{16cm}{\centering
    {\large Jiayi Wang\quad Franziska Mueller\quad Florian Bernard\quad Christian Theobalt}\\
    {\normalsize
    Max Planck Institute for Informatics, Saarbrücken, Germany\\}}
}

\begin{document}

\ifFGfinal
\thispagestyle{empty}
\pagestyle{empty}
\else
\author{Anonymous FG2020 submission\\ Paper ID \FGPaperID \\}
\pagestyle{plain}
\fi
\maketitle

%%%%%%%%%%%%%%%%%%%%%%%%%%%%%%%%%%%%%%%%%%%%%%%%%%%%%%%%%%%%%%%%%%%%%%%%%%%%%%%%
%%%%%%%%% ABSTRACT
\begin{abstract}
    We propose to use a model-based generative loss for training hand pose estimators on depth images based on a volumetric hand model.
    This additional loss allows training of a hand pose estimator that accurately infers the entire set of $21$ hand keypoints while only using supervision for $6$ easy-to-annotate keypoints (fingertips and wrist).
    We show that our partially-supervised method achieves results that are comparable to those of fully-supervised methods which enforce articulation consistency.
    Moreover, for the first time we demonstrate that such an approach can be used to train on datasets that have erroneous annotations, i.e.~``ground truth'' with notable measurement errors, while obtaining predictions that explain the depth images better than the given ``ground truth''.
\end{abstract}

%%%%%%%%% BODY TEXT
\section{Introduction}

Accurate hand-pose estimation from monocular depth images is vital for applications such as fine-grained control in human--computer interaction, or virtual and augmented reality~\cite{SolimanMHRTS18}.
However, it is a challenging task due to e.g.\,complex poses, self-similarities, and self-occlusions.
Many existing methods address these challenges with powerful learning-based tools.
Such methods dominate the benchmarks on large public datasets such as NYU~\cite{Tompson2014}, and Hands in the Million Challenge (HIM)~\cite{Yuan_2017_CVPR}.
Most of these approaches are trained in a fully supervised manner to predict the full set of 21 hand keypoint positions in 3D. 
However, the current lack of large-scale training datasets that are accurate and diverse causes such methods to overfit.
This makes it difficult to generalize well to new settings, or even across benchmarks~\cite{Yuan_2017_CVPR}.
Retraining these methods on different data requires the full set of $21$ (3D) keypoint annotations, which are tedious to obtain.
More importantly, this process is prone to errors in the data annotations, either due to measurement errors, or due to human errors.
Additionally, methods that learn a direct mapping from depth image to keypoints often ignore the inherent geometry of the hands, such as constant bone lengths or joint angle limits. 
As such, albeit their general good performance, these methods may produce bio-mechanically implausible poses~\cite{Wohlke2018}.
\begin{figure}[t]
\begin{center}
\includegraphics[width=1.0\linewidth]{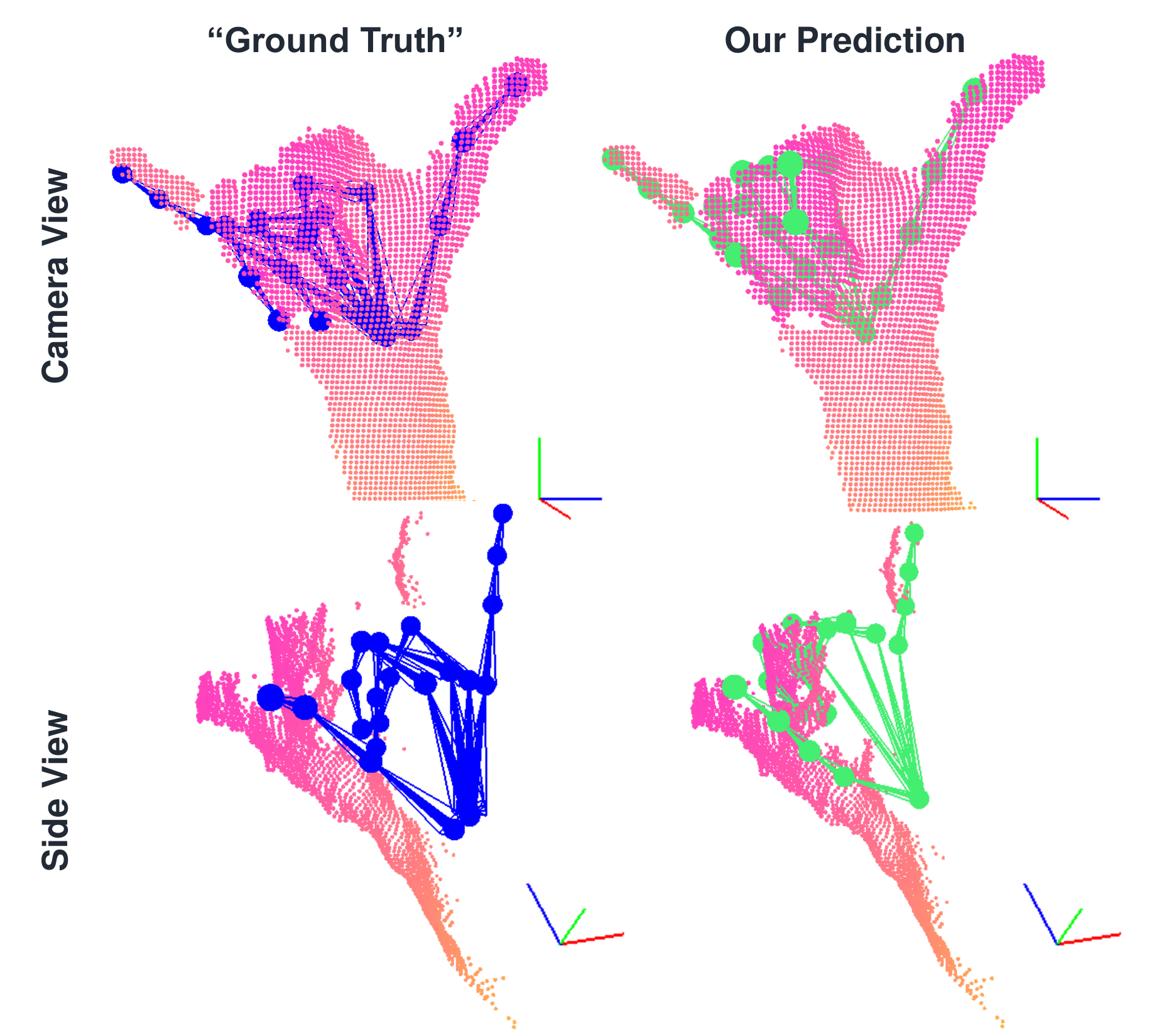}
\end{center}
    \vspace{-3.0mm}
   \caption{Our method uses self-supervision to compensate for erroneous ``ground truths'' (Blue), resulting in predictions (Green) that better fit the observed depth image.}
   \vspace{-7.0mm}
\label{fig:Teaser}
\end{figure}
An alternative to learning-based approaches are model-based hand tracking methods, such as~\cite{Melax:2013:DBS:2532129.2532141,FastHandTracker_CVPR2015,Taylor2017ArticulatedSDT,Tkach:2017:OGM:3130800.3130830}, among others.
These methods use generative hand models to recover the pose that best explains the image through an analysis-by-synthesis strategy. 
While not suffering from anatomical inconsistencies, and generalizing better to yet-unseen scenarios, they require good initialization of the model parameters in order to minimize the non-convex energy function. 

Our method addresses the shortcomings of both approaches with a generative model-based loss embedded into a learning-based method.
Based on a volumetric Gaussian hand model, this loss incorporates additional annotation-free self-supervision from the depth image. 
When combined with anatomical priors, this supervision can take the place of the majority of joint annotations for resolving hand pose and bone length ambiguities. 
In total, our approach reduces the number of required annotations  from $21$ to $6$, a $71$\% decrease. 
At the same time, the learning-based framework enables accurate and efficient inference during test time without requiring initialization.
This effectively combines the main advantages of the two popular categories.

Most existing methods that utilize a model-based loss ~\cite{Malik3DV2017, MalikEuroVR2017, Wohlke2018, Zhou:2016:MDH:3060832.3060960} do not explain the input images in a generative manner.
As such, they still require the full set of $21$ annotated keypoints per frame.
Additionally, due to the reliance on the annotations as the only source of supervision, these methods can overfit to errors and biases in the annotations.
We demonstrate that our method can overcome such errors through the use of our additional generative loss. 

We summarize our main contributions as follows:
\begin{itemize}[noitemsep,topsep=0pt,parsep=0pt,partopsep=0pt]
    \item Compared to classical fully supervised methods, our generative loss significantly reduces the amount of annotations need to accurately infer the full hand pose. 
    \item Despite ambiguities resulting from the reduced annotations, our method can simultaneously infer pose and bone lengths at test time. 
    \item We provide a new dataset, \textsc{HandId}, which includes fingertips and wrist annotations for $7$ users to address the lack of hand shape variations in existing datasets. 
    \item Most importantly, for the first time we demonstrate that such an approach can produce hand pose predictions that better fit to the depth image than the ``ground truth'' annotations it is trained on.
\end{itemize}
\section{Related Work}

Existing approaches for hand pose estimation can be broadly categorized into learning-based approaches, model-based approaches, and hybrid approaches.

\textbf{Discriminative, learning-based approaches.} These methods regress the pose parameters directly from image and annotation pairings.
Tompson et~al.~\cite{Tompson2014} first used a Convolutional Neural Network (CNN) for the task of hand pose estimation. 
From this foundation, many methods~\cite{OberwegerDeepPrior2015,DeepHand} develop novel architectures and training procedures to better model the nonlinear manifold of hand poses.
Recent methods investigate the use of different input representations such as multi-view, voxels, and point clouds, ~\cite{Ge2016Robust3H,3DCNN,Ge_2018_ECCV} to take advantage of known camera intrinsics.

\textbf{Generative, model-based approaches.} These methods iteratively refine an estimated pose by fitting a 3D hand model to the input depth image.
Previous work demonstrated that energies based on articulated, rigid, part-based models of the hand can be optimized to provide good tracking~\cite{Oikonomidis2011, Melax:2013:DBS:2532129.2532141}. 
Additional 3D hand representations, including continuous subdivision surfaces~\cite{TaylorContinuous}, collection of Gaussians~\cite{ellipsoidtracker_3dv2014, Stoll_2011_ICCV}, sphere meshes~\cite{TkachSphereMesh}, and articulated signed distance functions~\cite{Taylor2017ArticulatedSDT}, have been proposed with the goal of creating detailed models that are still fast to optimize.

\textbf{Hybrid approaches.} 
These methods combine learning-based and model-based approaches into one framework to combine the strengths of both. 
One class of hybrid methods uses learning-based components in a tracking framework to initialize, update, or otherwise guide the tracker's convergence to the correct pose~\cite{Oberweger:2015:TFL:2919332.2919832, Sharp:2015:ARF:2702123.2702179, FastHandTracker_CVPR2015, SunCascaded, TangICCV2015, Madadi_2018_IVC}.
These methods are more robust than the traditional model-based trackers, but must trade-off model and solver efficiency with accuracy during runtime.
Another class of hybrid methods uses the learning-based framework and incorporates a model-based loss, usually based on a kinematic skeleton ~\cite{Malik3DV2017, MalikEuroVR2017, Wohlke2018, Ye_ECCV_2016, Zhou:2016:MDH:3060832.3060960}.
These methods can better enforce anatomically plausible pose predictions by including pose priors losses in the model space. 
However, since the model is not generative, they still rely on difficult-to-acquire annotations and overfit to annotation errors if present.

Our proposed hybrid method incorporates a loss that is both \textit{generative} and \textit{model-based}, into the learning framework.
Unlike other hybrid approaches, the generative model provides supervision from the input depth image. 
With that, we are able to reduce the requirements on the quantity and accuracy of annotations needed for training, thereby reducing the necessary human effort for data annotation.

\begin{figure*}[th]
\begin{center}
\includegraphics[width=1.0\linewidth]{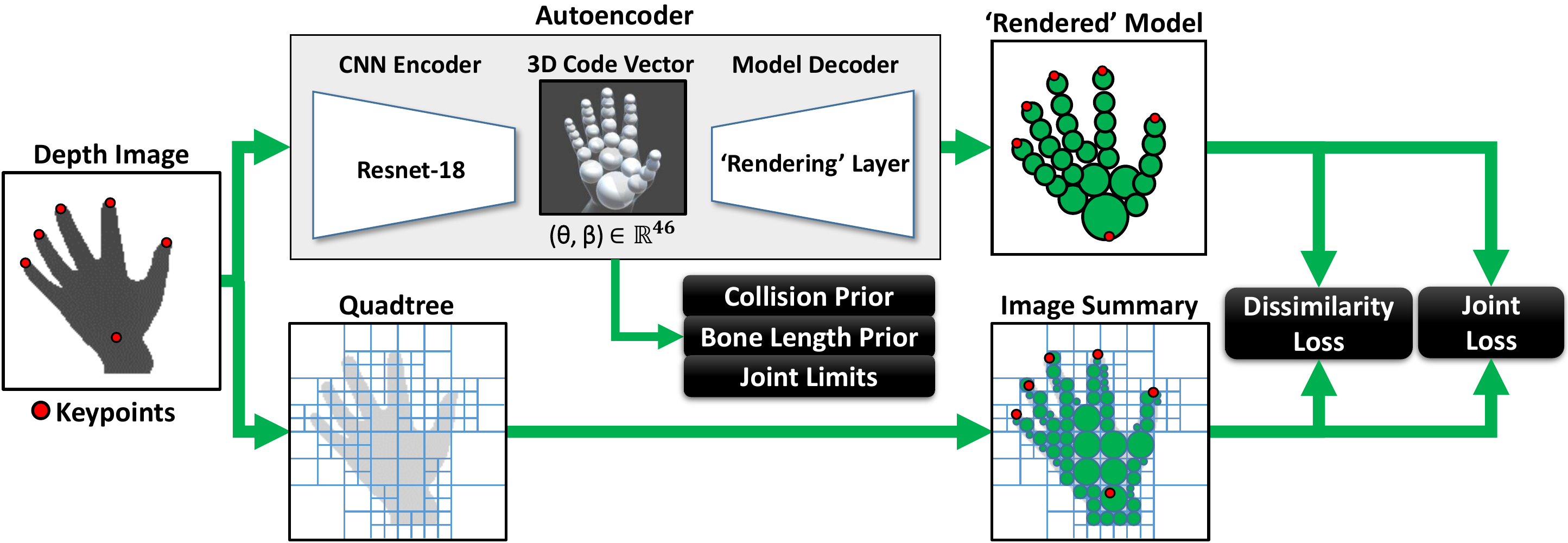}
\end{center}
    \vspace{-0.1cm}
   \caption{
   \textbf{Framework Overview.} 
   During training, an encoder is used to regress a code vector that parameterizes the bone lengths and pose in 3D. 
   A model-based generative decoder ``renders'' the 3D volumetric Gaussian hand into Gaussians in the image space. The original depth image is also summarized as Gaussians in image space through a Quadtree encoding.
   The dissimilarity between the two sets of Gaussians provides an unsupervised generative loss for training the encoder.
   Additionally, bone lengths and pose prior losses are used to regularize the encoding, and a partial supervision defined on a subset of the keypoints helps to overcome bad local optima in the dissimiliarity loss.
   }
\label{fig:pipeline}
\vspace{-0.45cm}
\end{figure*}

\parahead{Model-based Autoencoder.} 
Autoencoders are used for obtaining compressed representations from a distribution of inputs. 
They consist of an encoder that maps the input to a compact code, and a decoder that maps the code back to the (approximate) input.
Although the encoder and decoder are usually trained jointly, the encoder can learn to invert a generative model being used as the decoder in an self-supervised manner~\cite{NairInvertGenerativeBlackBoxes}.
As a learning objective, the model-based decoder can draw upon the entire training corpus as regularizer to overcome local minima that arise from noise or ambiguities present in a single image.
Tewari et~al.~\cite{Tewari_2017_ICCV} use such an autoencoder with a face model to estimate and disentangle face shape, expression, reflectance, and illumination. 
Recently, such approaches have also been proposed  for hand pose estimation in RGB images~\cite{Baek_2019_CVPR, Boukhayma_2019_CVPR, ge2019handshapepose}. These methods have in common that they use geometric cues (e.g.~annotated silhouettes and paired depth map) as supervision for training.
Dibra et~al.~\cite{Dibra_3DV} and Wan et~al.~\cite{Wan_2019_CVPR} use autoencoders for inverting a hand model to solve the hand pose estimation problem from depth images without additional cues. In contrast to~\cite{Dibra_3DV}, our use of a volumetric Gaussian hand model ~\cite{FastHandTracker_CVPR2015} as a decoder provides a stronger shape prior than their unconstrained articulating point cloud. This allows our method to solve the much harder problem of combined pose and shape estimation, while their method cannot adapt the hand shape at test time.
Although conceptually our method has similarities with the (concurrently developed) work~\cite{Wan_2019_CVPR}, our method uses a smooth hand representation compared to their spherical representation. More importantly, we extensively study the effect of a model-based generative loss when training with erroneous annotations (e.g.~as present in the HIM~\cite{Yuan_2017_CVPR} dataset), and hence we believe both works can be seen as complementary.

\section{Method}

The main idea of our approach is to explain a depth image of a hand based on a generative hand model, cf.~Fig.~\ref{fig:pipeline}.
Given a depth image as input, we use a CNN-based encoder to obtain a low-dimensional embedding of the depth image.
Our parametric model-based decoder is build upon a parametric hand model that produces a volumetric representation of the hand from a given code vector.
Since the code vector from the encoder initializes a parametric model, this enforces a semantically meaningful code vector. 
By using a suitable representation of the input depth image, we are able to efficiently and analytically compute the overlap between the ``rendered'' volumetric hand representation generated by the decoder and the input depth image.
To be more specific, we approximate the surface of the hand with a collection of 3D Gaussians rigidly attached to a kinematic hand skeleton model.
The corresponding Gaussians in image space can be obtained by projecting the 3D Gaussians using the camera intrinsics.
Moreover, the depth image is also represented with image space Gaussians by quadtree-decomposing the image into regions of homogeneous depth and fitting an image Gaussian to each region.
The similarity between the model and the image can then be described as the depth-weighted overlap of all pairs of model and image Gaussians.
This overlap serves as generative model-based loss during network training and ensures that the predicted hand faithfully represents the observed data.
To enforce plausible poses and bone lengths, we add additional prior losses to avoid inter-penetrations of hand parts, violations of joint limits, and unphysiological combinations of bone lengths. 
Lastly, supervision for a small subset of keypoints is provided as a way to mitigate the multiple minima present in the non-convex energy.
At test time, the so-trained encoder is able to directly regress the hand pose and bone length parameters.

\subsection{Hand Model}

\begin{figure}[t]
    \begin{center}
        \includegraphics[width=1.0\linewidth]{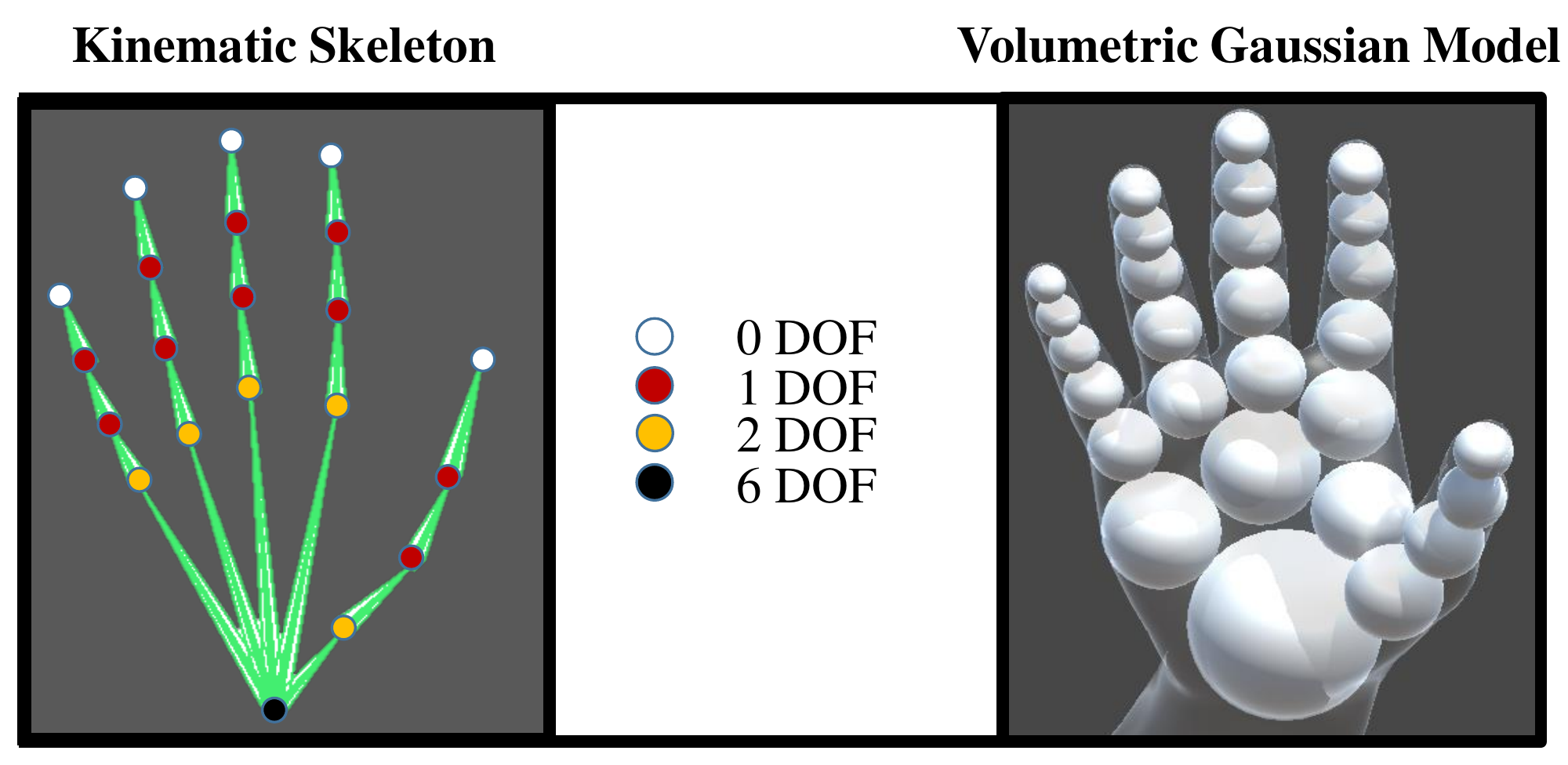}
    \end{center}
    \vspace{-0.3cm}
   \caption{ 
   \textbf{Left:} Our skeleton which comprises 20 bones and 15 articulating joints with varying degrees of freedom (DOF).
   In total, there are 26 joint parameters, and 20 bone length parameters.
   \textbf{Right}: Our volumetric Gaussian model.} 
\label{fig:kine_skel}
\vspace{-0.45cm}
\end{figure}

\parahead{Kinematic Skeleton.}
Our kinematic skeleton parameterizes hand shape in terms of bone lengths, and pose as articulation angles with respect to the joint axes.
It comprises 20 bones with lengths $b \in \mathbb{R}^{20} $ and 26 degrees of freedom (DOF) $\theta \in \mathbb{R}^{26}$ (20 angles of articulation and 6 additional DOF for global rotation and translation), see Fig.~\ref{fig:kine_skel}.

To ensure that the predicted bone length vector is plausible, $b$ is parameterized by an affine model constructed using 20 PCA basis vectors, i.e.
\begin{equation}
  \label{eq:PCA_Basis}
  b = b_{\text{avg}} + M_{\text{pca}} \beta \quad .
\end{equation}
	
Here, $b_{\text{avg}} \in \mathbb{R}^{20}$ is the average bone length vector and $M_{\text{pca}} \in \mathbb{R}^{20 \times 20}$ are the linear PCA basis vectors of the bone length variations scaled by their standard deviations. 
By scaling the basis vectors, $\beta$ follows an isotropic standard normal distribution, and deviations along each basis are penalized inversely to how much natural variation exists in that direction. Both $b_{\text{avg}}$ and $M_{\text{pca}}$ are obtained from bone length vectors computed from 10,000 hand meshes sampled from the linear PCA parameters of the MANO model~\cite{MANO:SIGGRAPHASIA:2017}.

The pose parameter vector $\theta$ controls the angles of articulation with respect to the joint axes in the forward kinematics chain, as well as the global translation and rotation of the entire hand, where the latter is is parameterized using Euler angles. 
Given the bone length parameters $\beta$ and pose $\theta$, we can obtain the $N_j$  joint positions by applying forward kinematics $F(\theta,\beta) \in \mathbb{R}^{N_j \times 3}$.

\parahead{Volumetric Gaussian Model.}
Similar to \cite{FastHandTracker_CVPR2015,Stoll_2011_ICCV}, we model the hand volume with a mixture of $N_m$ 3D Gaussians, i.e.
\begin{equation}
  \label{eq:Gauss_Mixture}
  G_{3D}(x) = \sum_{h=1}^{N_m} g_{\mu_{h}(\theta,\beta),\sigma_h}(x) \,,
\end{equation}
where $g$ is an isotropic Gaussian with mean $\mu_{h}(\theta,\beta)$ and standard deviation $\sigma_h$. 
Each Gaussian is attached to a bone on the kinematic skeleton and articulates with that bone. 

\subsection{Depth Image Representation}

The depth image is represented by a collection of 2D image Gaussian and depth value pairs $\{(g_{\mu_{i},\sigma_i}(x),  z_i)\}_{i=1}^{N_i}$. 
Each Gaussian and depth value pair summarizes a roughly homogeneous region with a single depth.
To obtain these regions, we use quadtree clustering to recursively divide the image into sub-quadrants until the depth difference within each region is below a threshold $c$ (we used $c = 20$mm for our experiments).
The Gaussian $g_{\mu_{i},\sigma_i}(x)$, is chosen so that $\mu_{i}$ is the center and $\sigma_i$ is half the side length of the region.
The associated depth value $z_i$ is then the average depth value of the quadrant.

\subsection{Model-based Decoder}

To measure the quality of the predicted hand pose and bone length parameters for a given input depth image, we incorporate a decoder layer that ``renders'' the 3D model representation to a 2.5D representation similar to the image representation.
The camera-facing surface of the $h$-th 3D Gaussian is approximated by a projected 2D Gaussian 
$g_{\mu_{p},\sigma_p}(x) = \Pi_{K}(g_{\mu_{h},\sigma_h}(x)) $ using the intrinsic camera matrix $K$ and an associated depth value $z_p$.
For details please refer to the supplemental document.

\subsection{Loss Layer}

For training the network, the loss is decomposed into an unsupervised dissimilarity term $E_{\text{dissim}}$ for 
measuring the discrepancy between depth image and hand model, $E_{\text{collision}}$ to prevent self intersection, $E_{\text{bone}}$ for regularizing the bone length parameters $\beta$, $E_{\text{lim}}$ for regularizing the joint angles $\theta$, and a supervised $E_{\text{joint}}$ term for explaining the provided joint locations. The relative importance of each term is balanced with scaling factors $\lambda$. With that, the total energy reads

\begin{equation}
  \label{eq:OverallEq}
  \begin{aligned}
  	E(\theta,\beta) = & \lambda_{\text{dissim}} E_{\text{dissim}}(\theta,\beta) +  \lambda_{\text{collision}} E_{\text{collision}}(\theta,\beta) + \\
  &	\lambda_{\text{bone}} E_{\text{bone}}(\beta) + \lambda_{\text{lim}} E_{\text{lim}}(\theta) + \lambda_{\text{joint}} E_{\text{joint}}(\theta,\beta)\,.
  \end{aligned}
\end{equation}
In the following we describe the individual energy terms. 

\subsubsection{Dissimilarity Measure}

To measure the overall similarity between two given (2D Gaussian, depth) tuples, we weight the similarity $S_{i,p}$ between the two Gaussians by their distance in depth values $\Delta(i,p)$.
The pairwise similarity between image Gaussian $g_{\mu_{i},\sigma_i}$ and projected model Gaussian $g_{\mu_{p},\sigma_p}$ is defined using the integral over the product of the two functions. 
Since in our case the model Gaussian directly depends on the hand pose vector $\theta$ and bone length vector $\beta$, $S_{i,p}$ is a function of these parameters and is given by
\begin{equation}
  \label{eq:gausssimilarity}
  	S_{i,p}(\theta,\beta) = \int\limits_{\mathbb{R}^2} g_{\mu_{i},\sigma_i}(x) g_{\mu_{p}(\theta,\beta),\sigma_p}(x)\: dx \,.
\end{equation}

Since $S_{i,p}(\theta,\beta)$ only measures the 2D overlap of the two Gaussians, we weight it based on the depth difference
\begin{equation}
  \label{eq:delta}
	\Delta(i,p) = \\
    	\begin{cases}0, & \text{if}\ |z_i - z_p| \geq 2\sigma_h\\
        	1-\frac{|z_i - z_p|}{2\sigma_h}, & \text{if}\ |z_i - z_p| < 2\sigma_h
    	\end{cases}
    	\,,
\end{equation}
where $\sigma_h$ is the standard deviation of the unprojected Gaussian $g_{\mu_{h},\sigma_h}$ associated with $g_{\mu_{p},\sigma_p}$. 
This decreases the similarity score between two tuples whenever the depth values are far apart, and thereby forces the model to not only match the area of the hand in the depth image, but also the observed depth values.

The overall similarity $S_{\text{sim}}$ is defined as the sum over all possible pairings between the model and the image Gaussians, and is given by
\begin{align}
  \label{eq:spatialalign}
	S_{\text{sim}} &= \frac{\sum_{i=1}^{N_i}\sum_{p=1}^{N_m} \Delta(i,p) S_{i,p}}{\sum_{i=1}^{N_i}\sum_{k=1}^{N_i} S_{i,k}} \quad,
\end{align}
where the denominator is the self-similarity of the image Gaussians used for normalization.
We use $E_{\text{dissim}} = -S_{\text{sim}}$ since minimizing the loss maximizes the similarity.

\subsubsection{Collision Prior}

To ensure that the surface represented by the 1$\sigma$ isosurface of the 3D Gaussians does not (self-)interpenetrate, a repulsive term based on the 3D overlap of the model Gaussians is used. 
Overloading the notation for the Gaussian overlap $S_{i,j}$ (cf.~Eq.~\eqref{eq:gausssimilarity}) to denote the similarity between two different model Gaussian components, we analogously define 

\begin{equation}
  \label{eq:collision}
  	E_{\text{collision}} = \sum_{j=1}^{N_m} \sum_{k=j+1}^{N_m} S_{j,k} \,,
\end{equation}
so that Gaussians of the model do not overlap in 3D. 

\subsubsection{Bone Length Prior}

To keep the bone lengths $\beta$ plausible, we impose the loss
\begin{equation}
  \label{eq:boneprior}
	E_{\text{bone}} = || \beta ||_2^2 \,,
\end{equation}
which penalizes the deviation of the predicted bone length parameters from the mean parameter. With that, this term helps to keep the predictions in the high probability region of the normal distribution used in the PCA prior. 

\subsubsection{Joint Limits}

To keep joint articulations within mechanically and anatomically plausible limits, a joint limit penalty is imposed using

\begin{equation}
  \label{eq:jointlimit}
	E_{\text{lim}} = \sum_{\theta_j \in \theta}\\
    	\begin{cases}0, & \text{if}\ \theta_j^l \leq \theta_j \leq \theta_j^h\\
        	(\theta_j^l - \theta_j)^2, & \text{if}\ \theta_j < \theta_j^l\\
            (\theta_j - \theta_j^h)^2, & \text{if}\ \theta_j > \theta_j^h
    	\end{cases} \quad ,
\end{equation}
where $\theta_j^l$ and $\theta_j^h$ are the lower and upper limits of $\theta_j$, which are defined based on anatomical studies of the hand~\cite{HandKinematicLimit}.

\subsubsection{Joint Location Supervision}

We impose an additional supervision loss $E_{\text{joint}}$ on a small subset of joint positions $J_1,\ldots,J_{N_s}$ in order to help the optimizer converge to a good minimum in the overall generative loss function. 
We use a combination of 2D and 3D joint location supervisions (depending on availability). 
If for a given joint with index $j$ a full 3D supervision is provided, the distance $\Phi_{j}$ between the annotation $J_j \in \mathbb{R}^3$ and the model joint $F_j$ is given by their $\ell_2$ distance. 
If only 2D supervision is provided, $\Phi_{j}$ is the closest $\ell_2$ distance between $F_j$ and the ray $\overline{J}_j$ to which the annotation is  projected using the camera intrinsics.
Hence, $\Phi_{j}$ is defined as
\begin{equation}
  \label{eq:joint_modified}
	\Phi_{j} =  
	\begin{cases}
	|| F_j - \langle F_j,\overline{J}_j \rangle \overline{J}_j ||_{2} \:, & \text{if } J_j \in \mathbb{R}^2 \\
    || F_j - J_j ||_{2}\:, & \text{if } J_j \in \mathbb{R}^3 \\
    \end{cases} \quad ,
\end{equation}
where $F_j = F(\theta,\beta)_j$ is the $j$-th joint obtained from applying forward kinematics with the model parameters.

Due to inaccuracies in the annotation, the ground truth may conflict with the observed image.
Hence, we modify the joint loss to account for annotation uncertainty by introducing a ``slack'' radius  $s \in \mathbb{R_+}$ that models the expected uncertainty in millimeters. 
All predictions within this radius of the ground truth will not be penalized. 
This allows the encoder to be more robust to erroneous annotations. 
Together, the joint loss for the subset of $N_s$ joints $E_{\text{joint}}$ is defined as

\begin{equation}
  \label{eq:slackjoint}
	E_{\text{joint}} = \sum_{j=1}^{N_s} \begin{cases}0, & \text{if}\ \Phi_{j} \leq s\\
        	(\Phi_{j} - s)^2, & \text{if}\ \Phi_{j} > s\\
    	\end{cases} \quad .
\end{equation}

\section{Experiments}

We evaluate the impact of our generative model-based loss on pose accuracy and bone length consistency when trained with a reduced set of keypoints.
Additionally, we show qualitative results of our predictions and the erroneous ``ground truth'' on existing datasets to demonstrate the regularizing effect of our loss against annotation errors.

\subsection{Architecture and Training}
We use Resnet-18~\cite{resnet_He_2016_CVPR} pre-trained on ImageNet as our encoder, as it is fast to use and refine, and achieves good accuracy. 
The encoder is trained with the Adam optimizer~\cite{ADAM_ICLR2015}, using a learning rate of $10^{-5}$ and a batch size of $16$. 
Our pipeline runs in Caffe~\cite{jia2014caffe}, where we implemented the decoder and other losses as custom layers.
During training, a forward-backward pass with batch size $16$ takes $89$ms (for comparison: ResNet-50 architecture takes $100$ms). A forward pass at test time takes only $5$ms.

\begin{figure}[t]
\begin{center}
\includegraphics[width=1.0\linewidth]{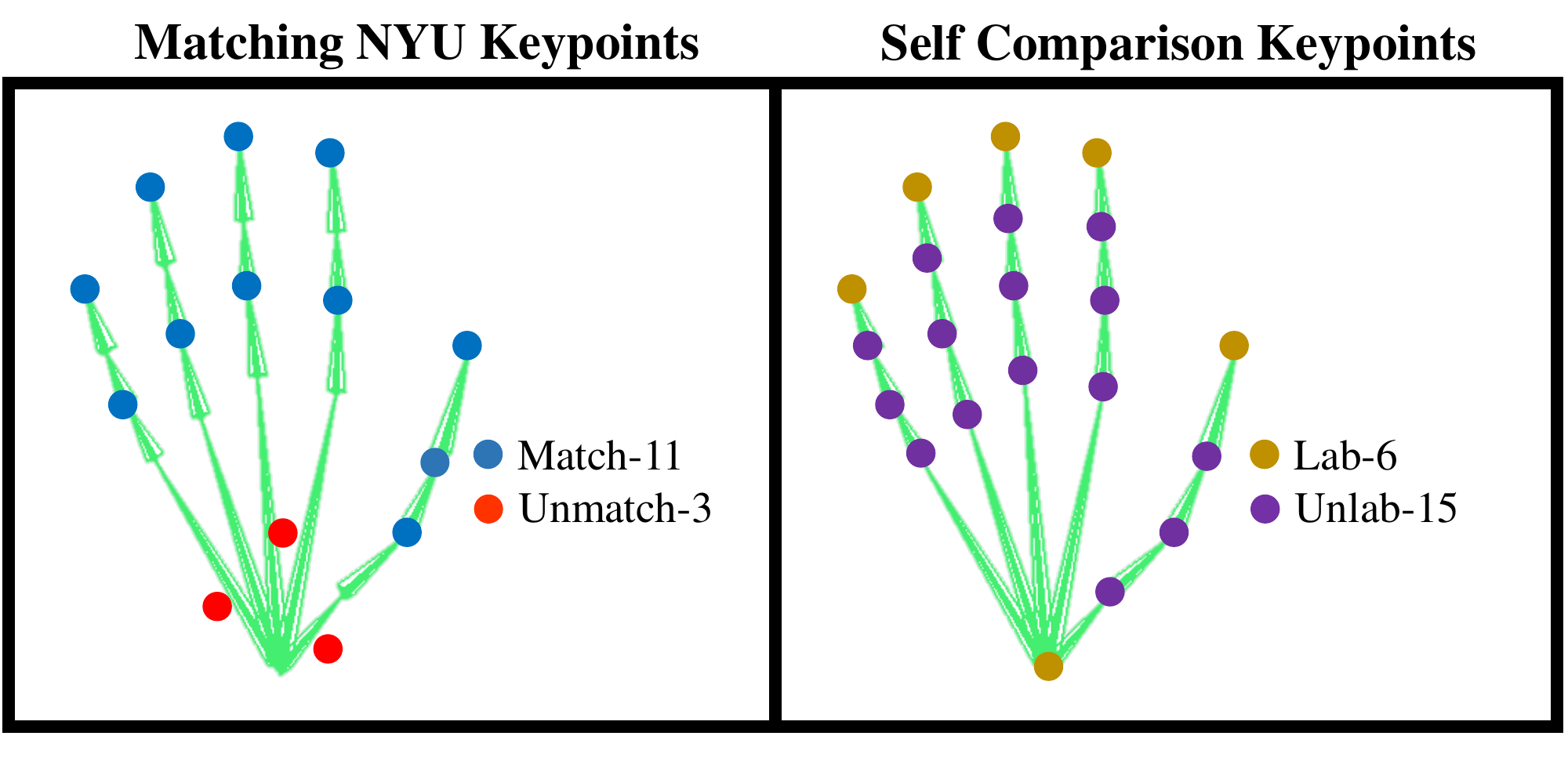}
\end{center}
    \vspace{-0.5cm}
   \caption{
    \textbf{Left:} For comparisons against the state of the art, our model is evaluated on a subset of NYU keypoints (\textbf{Match-11}) due to mismatches to our skeleton. 
    \textbf{Right:} For self-comparison, we evaluate on $21$ keypoints (\textbf{All-21}), $6$ of which have supervision (\textbf{Lab-6}), and $15$ keypoints without supervision (\textbf{Unlab-15}).
    }
\label{fig:ModelComparison}
\vspace{-0.45cm}
\end{figure}

\subsection{Datasets}
We evaluate on two common benchmarks, the NYU Hand Pose dataset~\cite{Tompson2014} and the Hands in the Million Challenge dataset (HIM)~\cite{yuan2018depth}. 
We additionally introduce our own \textsc{HandID} dataset for training to address the lack of hand shape variation in the NYU training data.

\begin{figure*}[t] 
   \begin{subfigure}[t]{0.32\textwidth}
\includegraphics[width=\columnwidth]{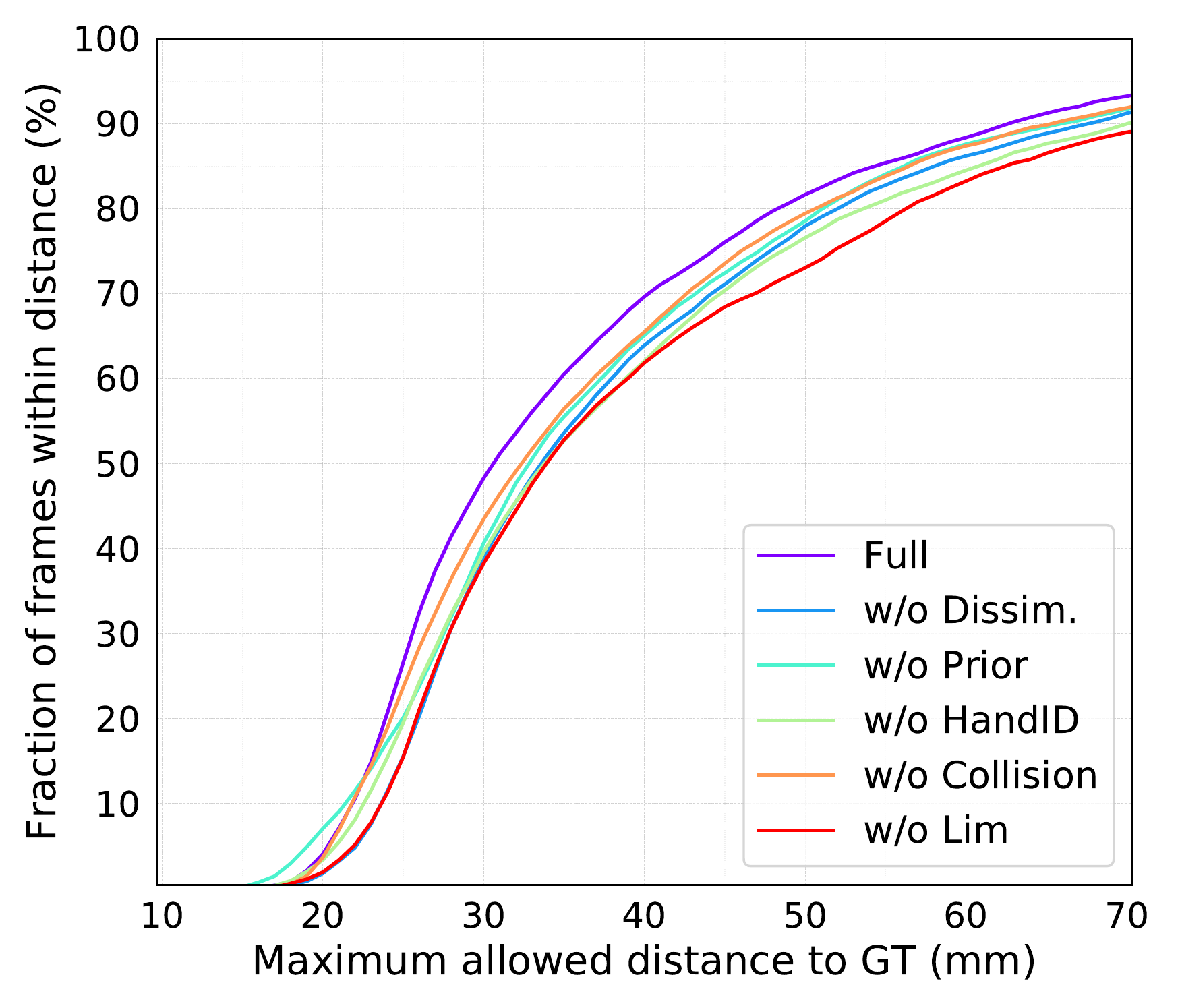}
   \caption{
   \textbf{Ablation Study:} All components of our method need to work together to resolve ambiguities from the reduced keypoint supervision (all keypoints (\textbf{All-21}) evaluated).}
\label{fig:curve_self}
	\end{subfigure}
	\hfill
    \begin{subfigure}[t]{0.32\textwidth}
\includegraphics[width=\columnwidth]{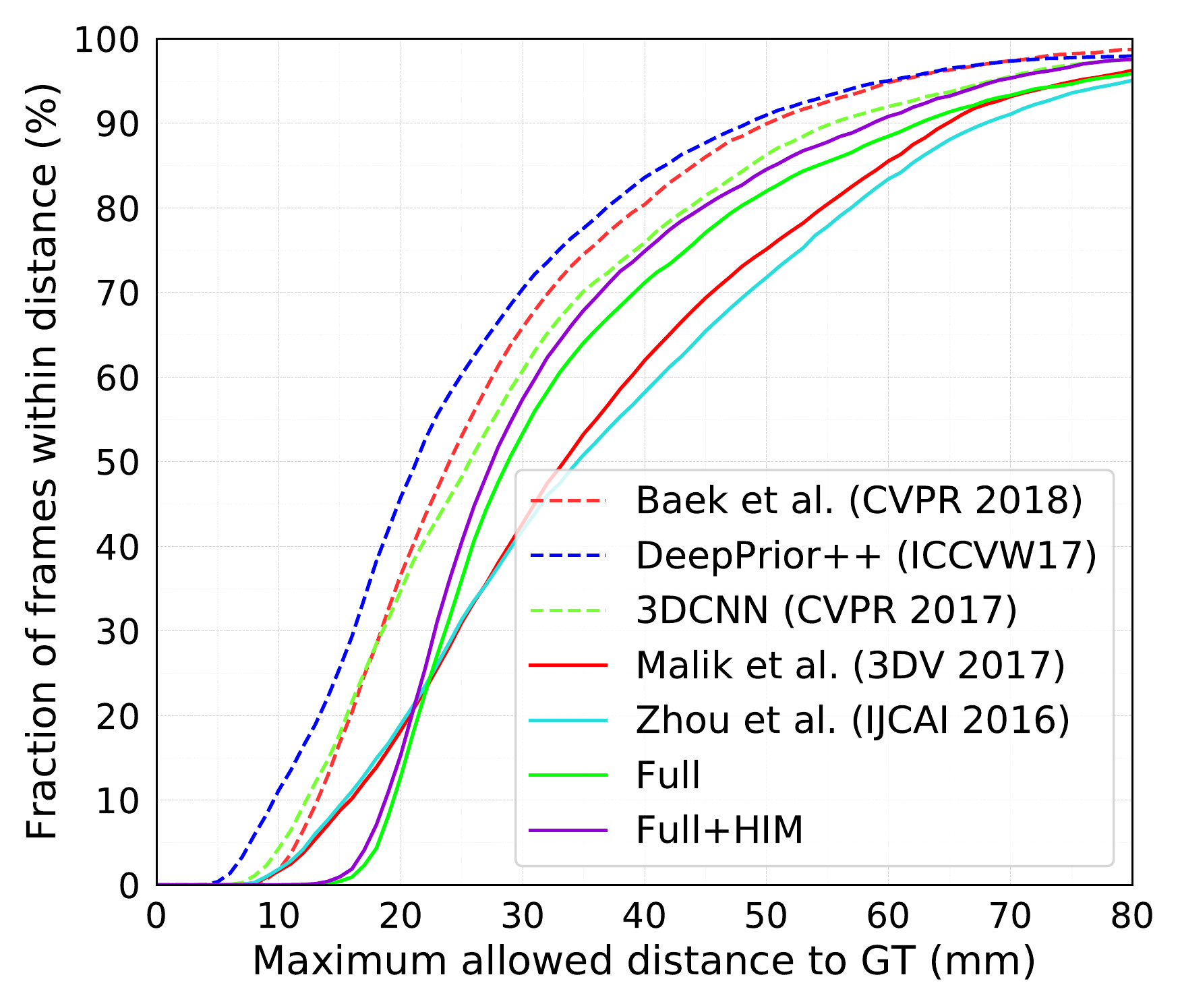}
   \caption{
   \textbf{Comparison to state of the art:} 
   Our method (\textbf{Full}) outperforms competing hybrid methods, even with less supervision. This is further improved by incorporating the HIM dataset, which is not possible without the dissimilarity loss.}
\label{fig:curve_STAR}
	\end{subfigure}
	\hfill
       \begin{subfigure}[t]{0.32\textwidth}
\includegraphics[width=\columnwidth]{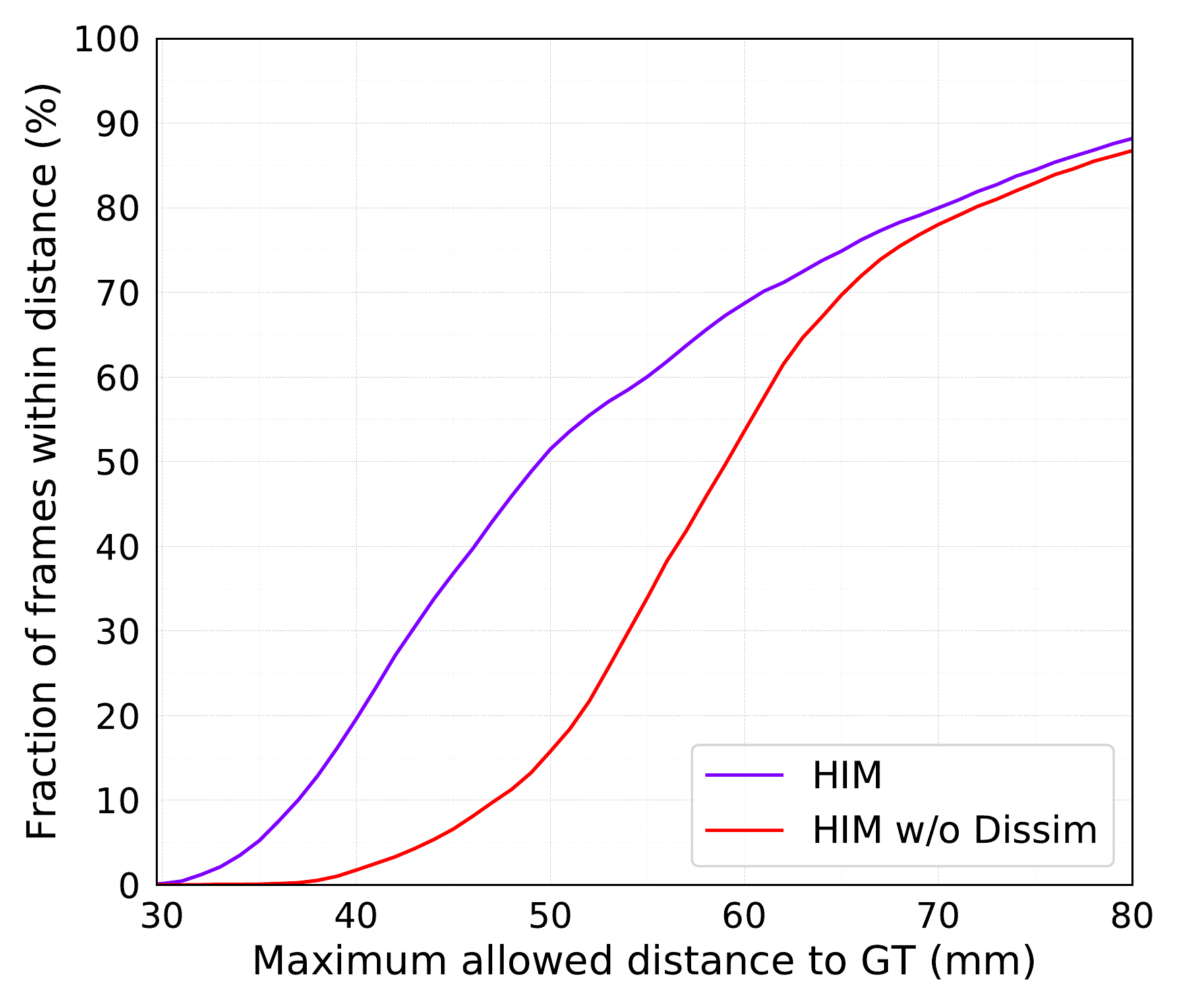}
   \caption{\textbf{Cross Benchmark Test:} We evaluate our method on the NYU dataset after training \emph{only on the HIM dataset}. Without the dissimilarity loss, the mismatch in annotation results in worse generalization.}
\label{fig:cross_benchmark}
	\end{subfigure}
  \caption{Quantitative evaluation on the NYU dataset (in percentage of frames with maximum joint error below a threshold).}
  \vspace{-0.3cm}
\end{figure*} 

\parahead{NYU Hand Pose Dataset.}
The NYU Hand Pose dataset~\cite{Tompson2014} is collected using Microsoft Kinect sensors. 
It contains $72{,}757$ depth images from a single subject in the training set, and $8{,}252$ depth images from two subjects in the test set. 

\parahead{Our \textsc{HandID} Dataset.}
Since the NYU training data only contains a single subject, we introduce additional training data with more hand shape variations to enable our method to learn this variation and hence adapt to different users at test time.
We captured a dataset of $3{,}601$ frames ($640$ x $480$) from $7$ subjects with the Intel SR300 sensor, which we call \textsc{HandID}.
A total of $6$ pixels that correspond to the fingertips and wrist are annotated per frame.
Occluded keypoints were indicated as such. 
During training, a batch contains examples from both \textsc{HandID} and the NYU dataset with a mixing ratio of $1:3$.

To emphasize that it is significantly easier to obtain just the fingertips and wrist keypoints, we asked $5$ users to annotate all $21$ keypoints for a set of $10$ depth images.
We observed that additional keypoints take longer to annotate (each joint annotation takes $1.4$ times longer) and are less consistent across users (with average distance to mean of $10.4$ pixels vs $7.3$ pixels).
In total, the full annotation of $21$ joints for $10$ images requires $21.2$ minutes, while our subset only needs $4.7$ minutes.

\begin{figure*}[t]
    \centering
    \textbf{Visualization of Annotation Errors}
      \vspace{2mm}
      \begin{subfigure}[b]{1.0\textwidth}
        \centering
          \includegraphics[width=\linewidth]{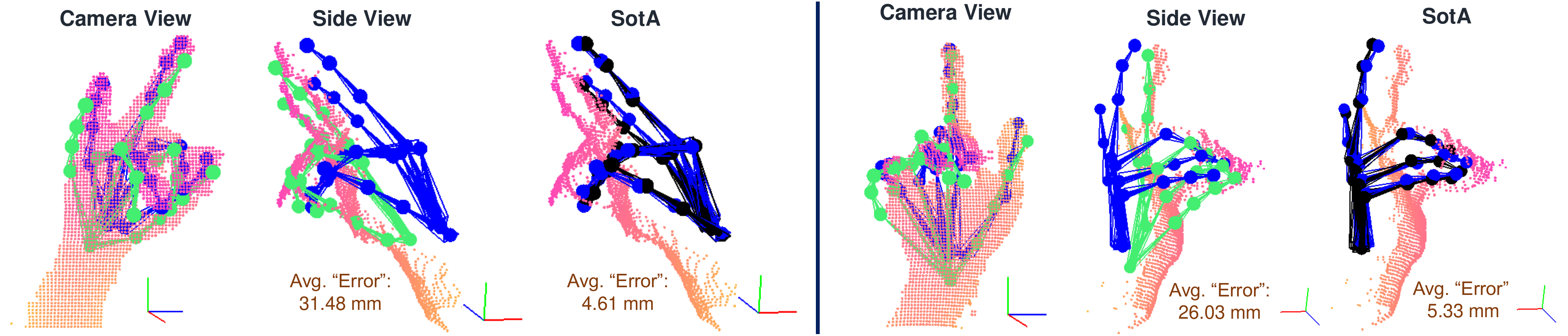}
             \label{fig:GoodAnnotation}
      \end{subfigure}
      \vspace{-4mm}
      
    \caption{\textbf{Annotation Errors in HIM:} Both the ``ground truth'' (Blue) and our predictions (Green) are consistent with the input in the camera view. However, as can be seen from the side view, the ``ground truth'' is erroneous and our prediction is more consistent. State-of-the-art (SotA)  method~\cite{Wu_2018_ECCV} (black) learns to replicate the systematic error. This result is representative of the test set.}\label{fig:HIMBias}
    \vspace{-4mm}
\end{figure*}

\parahead{Hands in the Million Challenge (HIM) Dataset.}
We evaluated our method on the Hands in the Million Challenge (HIM) dataset~\cite{yuan2018depth}, where we discovered a systematic error in the ``ground truth'' annotations.
Although the 2D projection of the keypoints into the image plane looks plausible, the 3D keypoint locations do not match the anatomical locations of hand joints (see Fig.~\ref{fig:HIMBias}).
To quantitatively show this, we use the minimum-distance-to-point-cloud (MDPC) per joint to approximately quantify how well the joint predictions agree with the observed depth image.
The NYU annotations and the erroneous HIM annotations have median MDPCs of $9.10$mm (avg $10.99$mm) and $21.54$mm (avg $23.98$mm), respectively.
By assuming that the physical joint is located roughly at the center of the finger, the HIM annotations would imply an implausible finger thickness of ${\approx}43$mm, while the NYU annotations estimates a more reasonable thickness of ${\approx}18$mm.
We hypothesize that there is a systematic pose-dependent error in corresponding the 3D magnetic sensor positions to the depth camera coordinate (see Fig.~4 of the Supplementary Document).
Using our generative model-based loss, we are able to obtain predictions that are significantly more consistent with the observed depth images.
The detailed experiment is presented in Section~\ref{sec:SotA}.

\begin{table*}[t]
\begin{subtable}[t]{0.39\textwidth}
\begin{center}
    \begin{tabular}{|l|c|c|c|}
\hline
Method & Unlab-$15$ & Lab-$6$ & All-$21$\\
\hline\hline
Full & \textbf{16.13} & \textbf{20.72} & \textbf{17.45}\\
w/o Dissim. & 19.06 & 21.47 & 19.75\\ 
w/o Prior & 18.53 & 22.03 & 19.53\\
w/o \textsc{HandID} &  17.01 & 23.20 & 18.78\\
w/o Collision &  16.80 & 22.20 & 18.34\\
w/o Lim. &  18.72 & 22.24 & 19.73\\ 
\hline
\end{tabular}    
\caption{
\textbf{Ablation study} with keypoints (see Fig.~\ref{fig:ModelComparison}) of the NYU dataset~\cite{Tompson2014}. 
Dissimilarity loss, and the pose and shape priors help resolve ambiguities for unlabeled keypoints.
The \textsc{HandID} dataset helps on labeled keypoints by allowing adaptions to unseen users.
}
\label{table:ablation}
\end{center}
\end{subtable}
\hfill
\begin{subtable}[t]{0.32\textwidth}
\begin{center}
\begin{tabular}{|l|c|}
\hline
Method & Match-$11$  \\
\hline\hline
Full & 18.50\\
Full+HIM w/o Dissim.& 20.01\\
Full+HIM& \textbf{17.73}\\
\hline
Zhou et~al.~\cite{Zhou:2016:MDH:3060832.3060960}& 19.21 \\
Malik et~al.~\cite{Malik3DV2017}& 18.35 \\
\hline
Baek et~al.~\cite{Baek_2018_CVPR}& 14.71 \\
DeepPrior++~\cite{DeepPriorPP} & 13.10\\
3DCNN~\cite{3DCNN} & 15.09 \\
\hline
\end{tabular}
\caption{
\textbf{Comparison to state of the art methods:} kinematic model-based (top, middle) enforces kinematic consistency and direct joint position regression (bottom) do not.
}

\label{table:STARcomparison}
\end{center}
\end{subtable}
\hfill
\begin{subtable}[t]{0.26\textwidth}
\begin{center}
\begin{tabular}{|l|c|c|}
\hline
Method & S1 & S2 \\
\hline\hline

Ground Truth & 1.00 & 1.00\\
Full+HIM & \textbf{0.70} & \textbf{0.80}\\
Full & 0.63 & 0.70\\
w/o Dissim. & 0.57 & 0.59\\
w/o Prior & 0.52 & 0.42\\
w/o \textsc{HandID}. & 0.55 & 0.54\\
w/o Collision & 0.62 & 0.68\\
w/o Lim. & 0.6 & 0.42\\
\hline
\end{tabular}
\caption{
F1 score of k-means clustering of bone lengths vectors for the two subjects in the test set.}
\label{table:bonelearning}
\end{center}
\end{subtable}

\caption{Evaluations on NYU. (a-b) Comparisons of 3D mean per-joint error (in mm). (c) Evaluation of bone lengths learning.}

\vspace{-3mm}
\end{table*}

\parahead{Pre-processing.}
Similar to established procedures~\cite{Baek_2018_CVPR,Oberweger:2015:TFL:2919332.2919832}, we first localize the hand by using the ground truth joint locations and crop the image to a fixed-size cube with $300$mm side length.
Once localized, the image is re-cropped using the same cube, but centered at the average depth.
We then scale it to $128$ x $128$ with a scaled depth range between $[-1,1]$. 
During training, in-image-plane translations and rotations, as well as depth augmentations, are applied.
This pre-processing step is used for all datasets. 

\parahead{Model Mismatch.}
Due to different joint locations in the NYU hand model and ours, only $11$ of the commonly evaluated keypoints have a rough equivalence to our model (Fig.~\ref{fig:ModelComparison}, left).
Hence, we compare our predictions with the state-of-the-art predictions on this subset (\textbf{Match-11}).
To better demonstrate that our method can infer the positions of unsupervised keypoints, we evaluate our algorithm for self comparison on an expanded set of $21$ NYU keypoints (\textbf{All-21}) which roughly correspond to anatomical joints of our kinematic skeleton (Fig.~\ref{fig:ModelComparison}, right).
The results are further broken down for the $6$ supervised keypoints (\textbf{Lab-6}) and the $15$ unsupervised keypoints (\textbf{Unlab-15}).

\subsection{Ablation Studies}
For the ablation study, we perform quantitative evaluations on the NYU dataset.

\parahead{Keypoint Accuracy.}
Removing components from our full method (\textbf{Full}) reduces accuracy.
See Table~\ref{table:ablation} for the average per-joint error in millimeters, and Fig.~\ref{fig:curve_self} for the percentage of correct frames curve.

\parahead{Bone Lengths.}
For bone length evaluation, we cannot directly compare the ground-truth bone lengths to our predicted bone lengths due to the mismatch in model definitions (cf.~Fig.~\ref{fig:ModelComparison}, left).
Instead, we treat the $20$ bone lengths of the hand as a $20$-dimensional vector and use k-means clustering with $k=2$ to separate the bone length vectors of the two subjects in the test set of the NYU dataset.
In Table~\ref{table:bonelearning}, we show the F1 scores (defined as $\frac{2{\cdot}\text{precision}{\cdot}\text{recall}}{\text{precision}{+}\text{recall}}$) of the two clusters. 
k-means is meaningful for this task as clustering bone lengths of the annotations (\textbf{Ground Truth}) results in perfect F1 scores for both subjects. 
Note that poses with high self-occlusion result in depth images with very little information to help disambiguate hand shapes.
Thus, one cannot expect methods that perform per-frame estimation to attain a perfect F1 score from the given supervision.

\parahead{Discussion.}
Given the reduced supervision, it is ambiguous whether the loss is minimized by deforming the bone lengths or updating the hand pose.
Consequently, the method without bone length prior can arbitrarily distort the bone lengths as long as the fingertips are correctly estimated (\textbf{w/o Prior}, see Table~\ref{table:ablation}).
This results in a significant drop in accuracy for keypoints without direct supervision (\textbf{Unlab-15}).
Correspondingly, k-means clustering fails to find consistent clusters for the two subjects.

However, the bone length prior alone is not enough to resolve the ambiguity in hand shape.
A similar drop in accuracy on unsupervised keypoints (\textbf{Unlab-15}) occurs when the dissimilarity loss is removed (\textbf{w/o Dissim.}, see Table~\ref{table:ablation}). 
This is because statistically plausible bone lengths can still vary wildly to accommodate the fingertip annotations, without being constrained to explain the image. 
Pose priors in the form of joint limits (\textbf{w/o Lim.}) and collision prior (\textbf{w/o Collision}) additionally constrain the articulations, which improve the keypoint accuracy.

Due to the NYU training data containing only one hand shape, it is sufficient for the method to consistently regress this particular set of bone lengths when \textsc{HandID} is not present (\textbf{w/o \textsc{HandID}}, see Table~\ref{table:ablation}).
As a result, the method cannot learn to discriminate between hand shapes of different users, leading to F1 scores that are close to random.
Hence, for the unseen hand shape in the test set, the method cannot minimize the joint loss (see Eq.~\eqref{eq:joint_modified}) of the supervised keypoints, which leads to greatly reduced accuracy on supervised keypoints (\textbf{Lab-6}). 
This mode of failure can be accounted for if hand shape variations are present in the training data. 
The result of this can be seen in our full method (\textbf{Full}, see Table~\ref{table:ablation}).

\subsection{Comparison to the State of the Art (SotA) \label{sec:SotA}}

Although state-of-the-art methods obtain mean per-joint errors lower than $10$mm (e.g. \cite{3DCNN, Wu_2018_ECCV}) on the HIM dataset, we emphasize that this is against the erroneous ``ground truth''. We train our method using a ``slack'' radius of $25$ mm to account for the error and show better fitting pose predictions than even the ``ground truth'' (see Fig.~\ref{fig:HIMBias} and Fig. 4 of Supplemental Material for more qualitative evaluation).

For a more fair quantitative evaluation, we instead use minimum-distance-to-point-cloud (MDPC) to approximate how well the predictions fit the input.
On the HIM test set of \cite{Wu_2018_ECCV} comprising of $95{,}540$ images, our method achieves median MDPCs of $11.74$mm (avg $13.87$mm), while~\cite{Wu_2018_ECCV} achieves $21.97$mm (avg $24.16$mm).
Our predictions better match the NYU annotations with median MDPCs of $9.10$ mm (avg $10.99$ mm).
This suggests that our method better fits the observed input while most state-of-the-art methods learn to replicate the errors in the training data.

We further show that the dissimilarity loss helps to overcome annotation errors by testing the method trained on HIM data on the NYU data (See Fig.~\ref{fig:cross_benchmark}). Without the dissimilarity loss, the method performs significantly worse.

On the NYU dataset (see Table~\ref{table:STARcomparison} and Fig.~\ref{fig:curve_STAR}), our method outperforms the other kinematic model-based methods of Zhou et~al.~\cite{Zhou:2016:MDH:3060832.3060960} and Malik et~al.~\cite{Malik3DV2017} while requiring less keypoint annotations. Although methods that directly predict 3D joint positions perform better~\cite{Baek_2018_CVPR, 3DCNN, DeepPriorPP}, we emphasize that these methods without a model-based generative loss are liable to learning the annotation errors as shown.

We compare our method to Dibra et~al.~\cite{Dibra_3DV} and  Wan et~al.~\cite{Wan_2019_CVPR}.
Although we were unable to obtain their predictions on the subset of \textbf{Match-11} keypoints, we note that Dibra et~al.~\cite{Dibra_3DV} have a similar ``uncorrected'' percentage of correct frames curve on all $14$ keypoints to Zhou et~al.~\cite{Zhou:2016:MDH:3060832.3060960}, which we greatly outperform, and we achieve similar performance to Wan et~al.~\cite{Wan_2019_CVPR}'s method with single view training.

While their methods do not require any annotation, our method additionally solves the more ambiguous and harder problem of adapting to the hand shapes of the user during test time, while their methods can only fit to the average hand shape of the training data or to preset bone lengths.

\subsection{Adaptation to a New Domain}

Despite the aforementioned annotation errors, the HIM dataset contains a large variety of views, poses, and hand shapes that could be used to supplement the NYU training data to help improve generalization.
We show that our method can still benefit from data with erroneous annotations (see Table~\ref{table:STARcomparison} and Fig.~\ref{fig:curve_STAR}).
We trained our method by mixing the NYU, HIM, and \textsc{HandID} datasets in a single batch with a ratio of $3$:$3$:$2$.
When HIM data is used without the dissimilarity loss (\textbf{Full + HIM w/o Dissim.}), the annotation errors cause the overall performance to degrade.
With our dissimilarity loss enabled (\textbf{Full + HIM}), the self-supervision ignores the annotation errors and improves the results.
\section{Limitations \& Discussion}

Although our method outperforms other kinematic model-based methods, even with less annotations, there is still a gap to recent learning-based methods that regress 3D joint positions. 
However, these methods
\begin{itemize}[noitemsep,topsep=0pt,parsep=0pt,partopsep=0pt]
    \item are not explicitly penalized for producing anatomically implausible shapes due to the lack of an underlying kinematic hand model, and
    \item are prone to overfit to errors in the training annotations, as well as to errors in the annotation collection method.
\end{itemize}

Additionally, for poses with heavy self-occlusions, the monocular depth data is not sufficient to resolve ambiguities with the reduced annotation set used by our method.
Extra supervision, such as from temporal consistency, or from multi-view constraints (as done in \cite{Wan_2019_CVPR}), is needed to estimate the pose and shape in these cases. 

\section{Conclusion}

We have shown that a generative model-based loss can reduce the amount of supervision needed to learn both the pose and shape of hands.
This greatly reduces the amount of annotations needed to adapt a method to data obtained in a new domain. 
Furthermore, we show that the generative model-based loss helps to regularize against annotation errors, for example on the HIM dataset, while existing methods overfit to these errors. 
This demonstrates the importance of ensuring that the model predictions explain not only the annotations but also the image itself.

\section*{Acknowledgments}
This work was supported by the ERC Consolidator Grant 4DRepLy (770784).

{\small
\bibliographystyle{ieee}
\bibliography{egbib}
}

\end{document}